\documentclass{article}

     \PassOptionsToPackage{numbers, compress}{natbib}

     \usepackage[preprint]{neurips_2020}

\usepackage[utf8]{inputenc} 
\usepackage[T1]{fontenc}    
\usepackage{hyperref}       
\usepackage{url}            
\usepackage{booktabs}       
\usepackage{amsfonts}       
\usepackage{nicefrac}       
\usepackage{microtype}      

\usepackage{graphicx, amsmath, subfigure, color, multirow, booktabs}

\newcommand{\nn}{\nonumber}

\newcommand{\figref}[1]{Figure~\ref{#1}}
\newcommand{\tabref}[1]{Table~\ref{#1}}

\def\bp{\mathbf p}
\def\bx{\mathbf x}
\def\bX{\mathbf X}
\def\by{\mathbf y}
\def\bz{\mathbf z}
\def\BbbR{\mathbb R}
\def\BbbE{\mathbb E}

\title{Learning Robust Feature Representations for Scene Text Detection}
\author{Sihwan Kim \\
  BigData and AI Laboratory\\
  Hana Institute of Technology, Hana TI\\
  Seoul 06133, Republic of Korea \\
  \texttt{shnasc.kim@hanafn.com} 
\And 
Taejang Park\thanks{Corresponding author} \\
	BigData and AI Laboratory\\
	Hana Institute of Technology, Hana TI\\
	Seoul 06133, Republic of Korea \\
	\texttt{djang000@hanafn.com} 
}

\begin{document}

\maketitle

\begin{abstract}
  Scene text detection based on deep neural networks have progressed substantially over the past years. 
  However, previous state-of-the-art methods may still fall short when dealing with challenging public benchmarks because the performances of algorithm are determined by the robust features extraction and components in network architecture. 
  To address this issue, we will present a network architecture derived from the loss to maximize conditional log-likelihood by optimizing the lower bound with a proper approximate posterior that has
  shown impressive performance in several generative models. 
  In addition, by extending the layer of latent variables to multiple layers, the network is able to learn robust features on scale with no task-specific regularization or data augmentation. 
  We provide a detailed analysis and show the results on three public benchmark datasets to confirm the efficiency and reliability of the proposed algorithm. 
  In experiments, the proposed algorithm significantly outperforms state-of-the-art methods in terms of both recall and precision.
  Specifically, it achieves an H-mean of 95.12 and 96.78 on ICDAR 2011 and ICDAR 2013, respectively.
\end{abstract}

\section{Introduction}\label{introduction}
Scene text detection is one of the most prolific research areas from the computer vision community due to its wide range of applications in multilingual translation, document analysis, scene understanding, autonomous driving, etc. 
Although previous works have made remarkable progresses, it is still challenging problem because of its various text shapes and highly complicated backgrounds. Recently, the text detectors based on deep learning
\cite{craft,sstd,R2cnn,east,fots,seglink,mask-textspotter,mostd,pixellink,stella,textsnake,textboxes,textboxes++} beyond the traditional approaches \cite{fastext,NM-stroke,sft,CY-drt,dfts,ZZ-cascade} have shown promising performance. 
While the traditional approaches have been designed to manually extract features of the scene texts, the deep learning based text detectors are designed to learn the features that are deemed useful from the training data. 
However, instead of designing, the network architecture,
most of the text detectors based on convolutional neural network have 
commonly focused their attention into generating the ground truth and post-processing in order to obtain single text instances.

To address this issue, we will introduce a novel network architecture derived from the loss to maximize conditional log-likelihood for text detection, {\it i.e.} $\log p(\by|\bx)$ where $\by$ and $\bx$ are the ground truth and input images, respectively. 
It can be conducted by optimizing the lower bound with a proper approximate posterior. 
While the most successful generative models often use only a single layer of latent variables \cite{vae}, our proposed model uses multiple layers of latent variables to effectively capture meaningful representations at each layer similar to \cite{hvae,lvae}. In this process, our model can collect rich features in different resolutions which allow the model to have robustness for scales on the texts and objects in the backgrounds.
Moreover, we are inspired by the previously developed multi-scale feature fusion techniques \cite{BiFPN,fpn,panet,nas-fpn} that can effectively aggregate features generated at each layer of the backbone network. 
Among them, we adopt a weighted bi-directional feature pyramid network \cite{BiFPN} named BiFPN which shows impressive performance in object detection tasks.
Compared to existing methods, the proposed algorithm achieves significantly enhanced performance according to qualitative and quantitative experiments on the public benchmarks.
Specifically, our method achieves an H-mean of 95.12 on ICDAR 2011 \cite{icdar11} and 96.78 on ICDAR 2013 \cite{icdar13}, outperforming the previous state-of-the-art methods in terms of both recall and precision.

The contributions of this work are summarized as follow:
\begin{itemize}
	\item We adopt a multi-scale feature fusion module to extract robust representations on text instances.
	\item Our method proposes a novel loss scheme for detecting text instances in natural scene images.
	\item Our method significantly outperforms the previous state-of-the-art method in both recall and precision on ICDAR 2011 and ICDAR 2013 datasets.
\end{itemize}

\section{Related Work}\label{sec:related-work}
During the past few years, detecting text in the natural scene image has been widely studied in the computer vision area. Before the emergence of deep learning era, most of the text detectors are based on  
Connected Components Analysis \cite{fastext,NM-stroke,sft} or Sliding Window \cite{CY-drt,dfts,ZZ-cascade}. 
After the appearance of deep learning based methods, most of the text detectors have adopted deep learning and gradually become the mainstream. 
The text detectors which are to localize texts in images can be classified into regression-based or segmentation-based approaches. 

\textbf{\textsl{Regression-based text detectors}}: Inspired by the popular object detection frameworks, such as SSD \cite{ssd} and Faster R-CNN \cite{faster-rcnn}, which predict the candidates in the form of coordinates of bounding boxes, various text detectors have been proposed. TextBoxes \cite{textboxes} that have quite a similar architecture to SSD \cite{ssd} modified convolutional filters and anchor boxes to effectively handle the highly complicated text instances.
Rotation Region Proposal Networks (RRPN) \cite{rrpn} inspired by Faster R-CNN \cite{faster-rcnn} introduced a novel framework to fit oriented text shapes by proposing rotational RPN and RoI pooling. 

\textbf{\textsl{Segmentation-based text detectors}}: Another popular approach is to regard text detection as semantic segmentation problem. These methods generally give pixel-level output maps indicating the probability of text regions.
The previous text detectors \cite{craft,pixellink,textsnake,holistic,pse} corresponding to this approach mainly adopt FCN \cite{fcn} and modify it to predict candidate of text region at pixel-level. 
While the segmentation-based approaches have the advantage of being able to predict text regions directly, identifying the positive regions from the output map into single text instances is not trivial. For example, multiple text instances lying very close to each other may make it hard to separate each instance. To efficiently separate these text instances, most segmentation-based text detectors need to additionally predict for desired single text instances.

The methods described as above have carefully designed the loss functions in terms of the output characteristics and the post-processing methods to obtain single text instances.
Meanwhile, we derive the loss function in a new direction deviated from the conventional perspective as the previous methods. 
We focus on how to design the network architecture to reflect the loss function which is derived to maximize conditional log-likelihood.
Our approach can be applied to all segmentation-based networks with slight architecture modifications because it is independent of 
the output characteristic and post-processing. To verify this, we will apply this idea to the network architecture of CRAFT \cite{craft}, which achieved state-of-the-art from the ICDAR 2013 dataset, and demonstrate the reliability and performance of our method with three public benchmarks, including ICDAR 2013.

\section{Proposed method}\label{sec:proposed-method}
In this section, we describe the framework of our proposed model and analyze the loss function in detail. 
Our network is a fully-convolutional neural network that outputs 
dense, per-pixel score maps named \textit{region score} to localize each individual character and \textit{affinity score} to link the semantically nearby characters as proposed in CRAFT \cite{craft}. 

\subsection{The lower bound of conditional log-likelihood}

Let $\mathbf X = \{(\bx_j, \by_j)_{j=1}^N\}$ be the training image datasets, where $\bx_j$ is the input image and $\by_j$ represents both the region and affinity score maps, consisting of $N$ i.i.d. observations.
We introduce latent variables $\bz = \{\bz_1, \cdots, \bz_L\}$ where $\bz_1$ represents the layer
closest to $\mathbf x$ and $\bz_L$ represents the top layer. 
Extending a single layer of latent variables for abstract feature representations allows the model to utilize deeper and more distributed latent variables \cite{hvae,lvae}.
Hence, we carefully design a network architecture to learn features of different levels between the dataset $\bX$ and the latent variables $\bz$. 
In our case, we assume that the latent variables $\bz$ are independent for each variable $\bz_l$. We wish to maximize the conditional log-likelihood $\log p(\by|\bx)$ over a datasets $\bX$. 
Formally, we would like to maximize
\begin{eqnarray*}\label{eq:cond_prob}
	\log p(\by|\bx) = \sum_{j=1}^N\log p(\by_j|\bx_j),
\end{eqnarray*}
which is non-convex and often intractable in general. To avoid the intractable integral, we introduce an approximate posterior $q(\bz|\bx_j)$ to obtain the lower bound using Jensen's inequality and the fact that the integral of probability distribution equal to 1.

\begin{eqnarray}\label{eq:elbo}
\log p(\by_j|\bx_j) 
&\geq& \int \log \Bigg( p(\by_j | \bz, \bx_j) \frac{p(\bz| \bx_j)}{q(\bz|\bx_j)} \Bigg)q(\bz|\bx_j)~d{\bz}, \nn\\
&=& \int \log \Big( p(\by_j | \bz, \bx_j)\Big) q(\bz|\bx_j)~d{\bz} - \int \log \Bigg( \prod_{l=1}^{L}\frac{ q(\bz_l| \bx_j)}{p(\bz_l|\bx_j)} \Bigg)q(\bz|\bx_j)~d{\bz}, \nn \\
&=& \BbbE_{q(\bz|\bx_j)}\Big[ \log p(\by_j|\bz,\bx_j) \Big] - \sum_{l=1}^L D_{KL} \Big(q(\bz_l|\bx_j)~||~p(\bz_l|\bx_j)\Big),
\end{eqnarray}
where $D_{KL}$ denotes Kullback-Leibler divergence. The next sub-section describes the network architecture and loss function in detail.

\subsubsection{Network Architecture}

\begin{figure}[h]
	\centering
	\includegraphics[width=1.0\linewidth, height=0.55\linewidth]{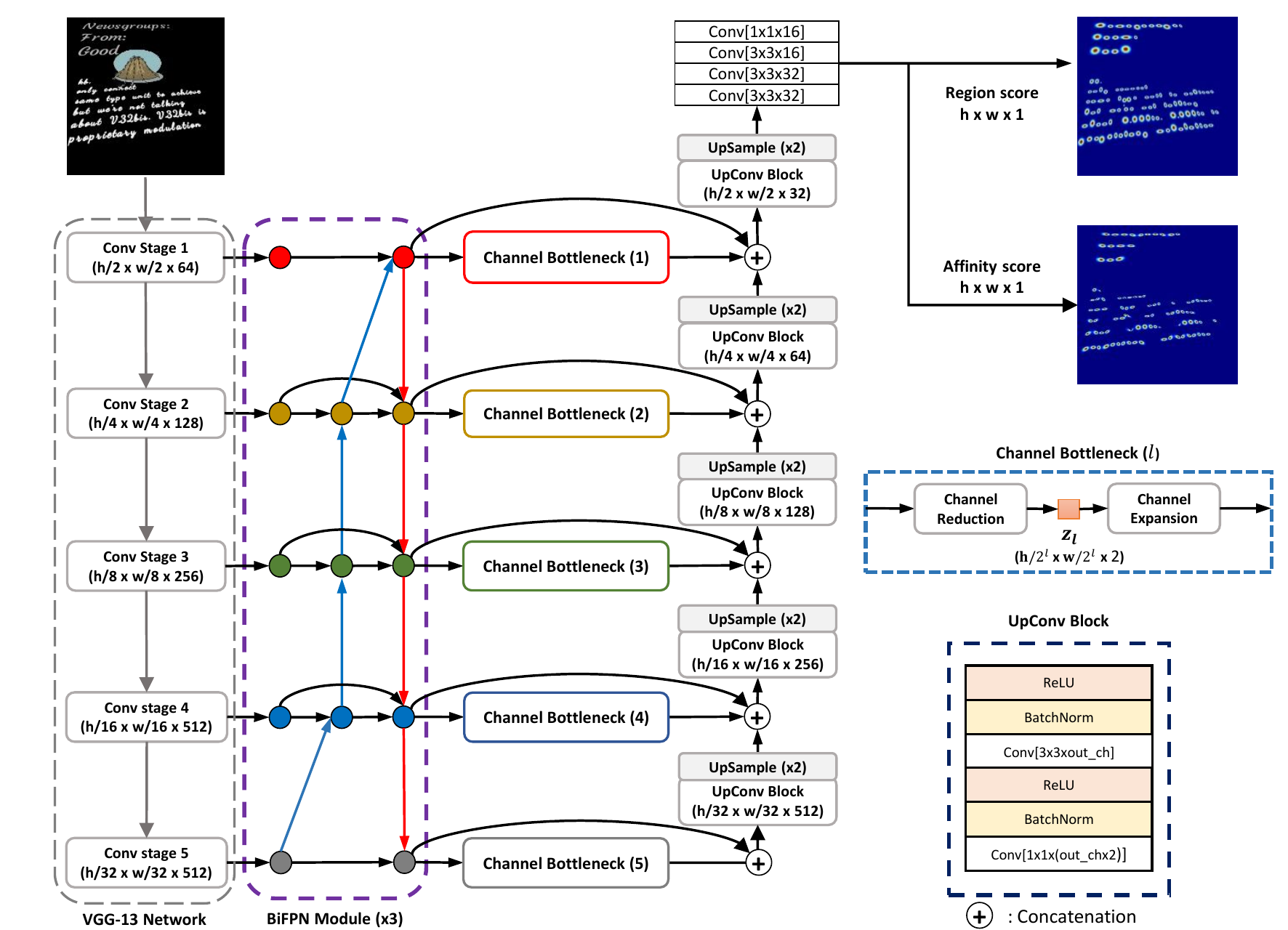}
	\caption{Network Architecture. The 13 convolutional layer inherited from VGG-16 \cite{vgg} with Batch normalization is adopted as the backbone network and then
		BiFPN module \cite{BiFPN} as the multi-scale feature fusion are repeated 3 times. After feeding channel bottlenecks to calculate KL-divergence term, 
		refine the outputs to make two score maps: the \textit{region score} and \textit{affinity score}}
	\label{fig:network}
\end{figure}

We adopt a fully convolutional network architecture based on VGG-16 \cite{vgg} with batch normalization except for the last three fully-connected layers as our backbone. 
In addition, we adopt the BiFPN modules that efficiently aggregates the multi-scale features with different resolutions to make a list of new features. 
Formally, given features $f^{in} = (f_{1}^{in}, \cdots, f_{5}^{in} )$ where $f_{l}^{in}$ represents the feature generated from level $l$ with resolution of $1/2^{l}$ of the input images. 
We would like to make new features $f^{out} = (f_{1}^{out}, \cdots, f_{5}^{out} )$ by a transformation $\mathcal{F}$ that can effectively aggregate multi-scale features.
A common technique when fusing multiple input features with different resolutions is to resize them to the same resolution and then sum all up with equal weights \cite{motd,scrdet}.
However, we intuitively believe that the input features with different resolutions could contribute to the output unequally.
To deal with this issue, the weighted feature fusion techniques have been developed in \cite{BiFPN}. Among them, we will adopt Fast Normalized Fusion technique that the weight $w_i \geq 0 $ are ensured by applying a ReLU to reduce computational cost while keeping the benefits of normalized weights.
For a concrete example, we describe the case of level 4 which is in blue circles in the \figref{fig:network}.
\begin{eqnarray*}
f_4^{td} &=& Conv\Bigg( \frac{w_1 f_4^{in} + w_2 Resize(f_5^{in})}{w_1+w_2+\epsilon} \Bigg), \\
f_4^{out} &=& Conv\Bigg( \frac{w_1^{'} f_4^{in} + w_2^{'} f_4^{td} + w_3^{'} Resize(f_3^{out})}{w_1^{'}+w_2^{'}+w_3^{'}+\epsilon} \Bigg),
\end{eqnarray*}
where $\epsilon = 0.0001$ is a small value to avoid numerical instability, $Resize$ is conventional upsampling or downsampling operation, $f_4^{td}$ and $f_4^{out}$ are the intermediate and output feature at level 4, respectively. $Conv$ implies Convolution-Batch normalization-ReLU operation consisting of $1\times1$ and $3\times3$ receptive fields with 128 and 64 filters, respectively.

The generated new features feed into channel bottlenecks similar to a standard auto-encoder that only transforms the dimension of a channel to extract latent variables $\bz_j, j=1,\cdots,5$. The reduction part have 3 layers with 32, 16 and 2 filters consisting of $1\times1$ and $3\times3$ receptive fields each. 
The expansion part consists of transformed structure of the reduction part.
The latter part of our network refines the outputs from the channel bottlenecks into two score maps: the \textit{region score} and the \textit{affinity score} to indicate the text regions and the relationships between semantically nearby characters, respectively. An overview of our network framework is illustrated in \figref{fig:network}.

\subsubsection{Loss Function}
We analyze the loss function described in \eqref{eq:elbo} in detail. 
The most critical point to be determined is how to assign a proper probability distribution to the latent space of the input data. Multivariate Gaussian distribution to real valued attributes or independent Bernoulli distribution to each binary attributes have been generally used in such cases. 

Let $F_{id}:\BbbR^D \rightarrow \BbbR^D$ be an identity function such that $F_{id}(\bx_j) = \by_j$. Also denote $F_{ip}^S:\BbbR^T \rightarrow \BbbR^S$ be an interpolation function by resampling using pixel area relations
to prevent aliasing phenomenon when a image is downsampled. 
In our case, the (conditional) prior $p(\bz_l|\bx_j)$ and the model $q(\bz_l|\bx_j)$ can both be modeled using Bernoulli distributions:
\begin{eqnarray*}
p(\bz_l|\bx_j) = \prod_{k=1}^{Q(l)}Ber\Big(\bz_l^k; F_{ip}^{Q(l)} \circ F_{id}(\bx_j) \Big), \qquad
q(\bz_l|\bx_j) = \prod_{k=1}^{Q(l)}Ber\Big(\bz_l^k; \theta^k(\bx_j) \Big), 
\end{eqnarray*}
where $Q(l)$ is the dimension of the latent variables at $l$-th layer and $\theta^k(\bx_j)$ are predicted by the model $q(\bz_l|\bx_j)$. 
For convenience, we use notation $\by_{j,l}$ to denote the mapping $F_{ip}^{Q(l)} \circ F_{id}(\bx_j)$.
The first term of \eqref{eq:elbo} can be calculated by the Stochastic Gradient Variational Bayes (SGVB) where the Monte Carlo estimates are performed with the reparameterization trick. For the Bernoulli distribution case, the reparameterization trick can be performed using a uniformly distributed auxiliary variable $\epsilon \sim U(\textbf{0},\textbf{1})$ described in \cite{JGP,MMT}.
A single sample is often sufficient to form the Monte Carlo estimates in practice, thus, we get
\begin{eqnarray}\label{eq:reconstruct}
\BbbE_{q(\bz|\bx_j)}\Big[ \log p(\by_j|\bz,\bx_j) \Big] = \sum_{k=1}^D \by_j^k \log \bp_j^k + (1-\by_j^k) \log ( 1 - \bp_j^k ),
\end{eqnarray}

where $\by_j^k$ and $\bp_j^k$ are $k$-th element of the ground truths and the estimated outputs, respectively.
KL divergence in the second term of \eqref{eq:elbo} is analytically computed as follow.
\begin{eqnarray}\label{eq:kl}
-\sum_{l=1}^L D_{KL} \Big(q(\bz_l|\bx_j)~||~p(\bz_l|\bx_j)\Big) &=& \sum_{l=1}^L \int q(\bz_l|\bx_j)\Big( \log p(\bz_l|\bx_j) - \log q(\bz_l|\bx_j) \Big)~d{\bz_l}, \nn \\
&=& \sum_{l=1}^L \BbbE_{q(\bz_l|\bx_j)}\Big[ \log p(\bz_l|\bx_j) \Big] - \BbbE_{q(\bz_l|\bx_j)}\Big[ \log q(\bz_l|\bx_j) \Big], \nn \\
&=& \sum_{l=1}^L \sum_{k=1}^{Q(l)} \bz_l^k \log \frac{\by_{j,l}^k}{\bz_l^k} + (1-\bz_l^k) \log \frac{ (1-\by_{j,l}^k)}{ (1-\bz_l^k)},
\end{eqnarray}
where $\bz_l^k$ and $\by_{j,l}^k$ are $k$-th element of the latent variables and the transformed value of $\bx_j$ by the mapping defined as above at $l$-th layer, respectively. 
By taking the sum of \eqref{eq:reconstruct} and \eqref{eq:kl}, we get

\begin{equation}\label{eq:loss}
\log p(\by_j|\bx_j) \geq \mathcal{L}_{recons} + \mathcal{L}_{reg},
\end{equation}
where $\mathcal{L}_{recons}$ and $\mathcal{L}_{reg}$ are reconstruction loss \eqref{eq:reconstruct} and regularization loss \eqref{eq:kl}, respectively.

Our proposed model will be trained to minimize the negative 
log-likelihood defined by \eqref{eq:loss}, which implies minimizing the reconstruction loss \eqref{eq:reconstruct} and the regularization loss \eqref{eq:kl}. The regularization loss encourages the approximate posterior $q(\bz_l|\bx_j)$ to be close to the Bernoulli distribution $p(\bz_l|\bx_j)$ defined by the resized ground truth with the shape corresponding to each $l$-th layer. 
Thus, due to the regularization loss, when the network is fully trained, the defined distribution at each $l$-th layer is naturally induced in our model.
This implies that our model is able to learn robust representations for scales on texts and objects in backgrounds. 

\subsection{Label Generation}

\begin{figure}[t]
	\centering
	\includegraphics[width=1.0\linewidth, height=0.3\linewidth]{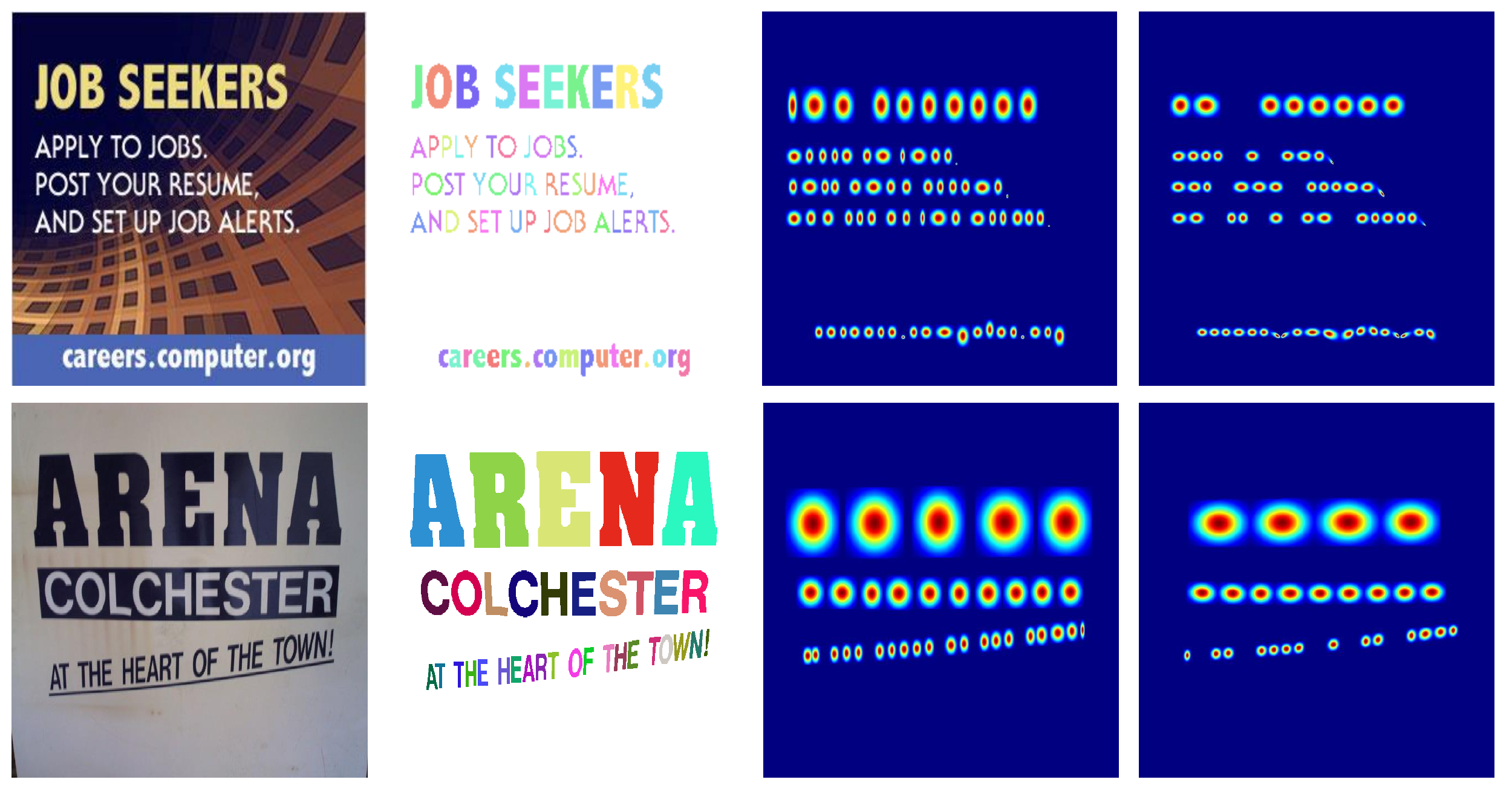}
	\caption{The examples of generated ground truth of ICDAR 2011 and ICDAR 2013. 1st column: Input image, 2nd column: The map for text segmentation, 3rd-4th columns: The generated \textit{region score} and \textit{affinity score}. }
	\label{fig:label_generation}
\end{figure}

For generating region and affinity score maps, which represent the probability of the center of the character and the relationships between semantically nearby characters in single text instances, respectively, we adopt the method of label generation described in CRAFT \cite{craft}.
Although ICDAR 2011 and ICDAR 2013 datasets do not provide character-level annotations directly, it is possible to generate ground truth labels by using extra information provided for text segmentation.
In the case of ICDAR 2011, the dataset for text segmentation is provided in the form of images where white pixels are interpreted as the background while 
non-white pixels are color coded so that pixels of each character in the image are shown in the same color. 
Hence the individual characters in word-level bounding box can be easily separated by the color identity. 
For ICDAR 2013, which is analogous to ICDAR 2011, the ground truth label can be also generated by separating characters in the images with the dataset provided for text segmentation. 
The examples of generated ground truth labels for the case of ICDAR 2011 and ICDAR 2013 are illustrated in \figref{fig:label_generation}.

\section{Experiments}\label{sec:experiments}
To verify the reliability of our proposed method, we used four public benchmarks, which are briefly introduced below, three of which are used to evaluate our method.

\subsection{Benchmark Datasets}
\textbf{SynthText} \cite{synth} is a large scale datasets that contains about 800K synthetic images. These images are synthetically created by blending rendered words with natural images.
This dataset is used for pre-training our model.

\textbf{ICDAR 2011} \cite{icdar11} is proposed in the Challenge 1 of the 2011 Robust Reading Competition for Web and Email scene text detection, consisting of 410 training images and 141 testing images. 
The text regions are annotated in the form of axis-aligned bounding boxes. 

\textbf{ICDAR 2013} \cite{icdar13} is a datasets proposed in the Challenge 2 of the 2013 Robust Reading Competition for focused horizontal text detection in scenes. 
It consists of 229 images and 233 images for training and testing, respectively.
The text regions are annotated in the form of axis-aligned bounding boxes. 

\textbf{ICDAR 2015} \cite{icdar15} is used in Challenge 4 of the 2015 Robust Reading Competition. It includes a total of 1500 scene images consisting of 1000 training images and 500 testing images
with annotations labeled as word level quadrangles.

\subsection{Implementation Details}
The network is pre-trained on SynthText for 50K iterations and fine-tuned it on both ICDAR 2011 and ICDAR 2013 together. The longer side of the training images are resized to 640, preserving their aspect ratio.
We adopt Adam optimizer \cite{adam} as our learning rate scheme. The learning rate is set to $10^{-3}$ initially and decays exponentially with a rate of 0.95 every 1K iterations on both pre-training and fine-tuning stage.
We train our model with the batch size of 32 on 8 GPUs in parallel, {\it i.e.} 4 images per GPU, and use synchronized batch normalization to effectively calculate the moving average and moving variance. 
To resolve the imbalance between positive and negative samples when calculating the loss defined by \eqref{eq:loss}, we use the Online Hard Negative Mining proposed in \cite{ohnm} and set the ratio of positives to negatives to 1:3.
Also, the data augmentation techniques used to train the model are: (1) random rotation with an angular range of -10 to 10; (2) random flipping with a probability of 0.5; (3) color-channel swapping with a probability of 0.5.

During the testing, the estimated \textit{region score} and \textit{affinity score} are binarized with the value $\tau_r$ and $\tau_a$
where $\tau_r$ and $\tau_a$ are the region and affinity threshold, respectively.
By taking the boolean addition of these binarized score maps, we get the final output map $M$. Then Connected Component Labeling (CCL) is performed on $M$ and single text instances are obtained by finding each quadrilateral bounding box enclosing the minimum area of each label.

\subsection{Experiment Results}
We evaluate the proposed method on three challenging public benchmarks: ICDAR 2011 \cite{icdar11}, ICDAR 2013 \cite{icdar13} and ICDAR 2015 \cite{icdar15}.
We use both ICDAR 2011 and ICDAR 2013 datasets for training, and ICDAR 2015 dataset is only used for testing as unseen data because ICDAR 2015 does not provide character-level annotations.
We analyze the results both qualitatively and quantitatively to confirm the efficiency and reliability of our proposed method.

\begin{figure}[t]
	\centering
	\subfigure [Correct examples] {\includegraphics[width=0.62\linewidth, height=0.33\linewidth]{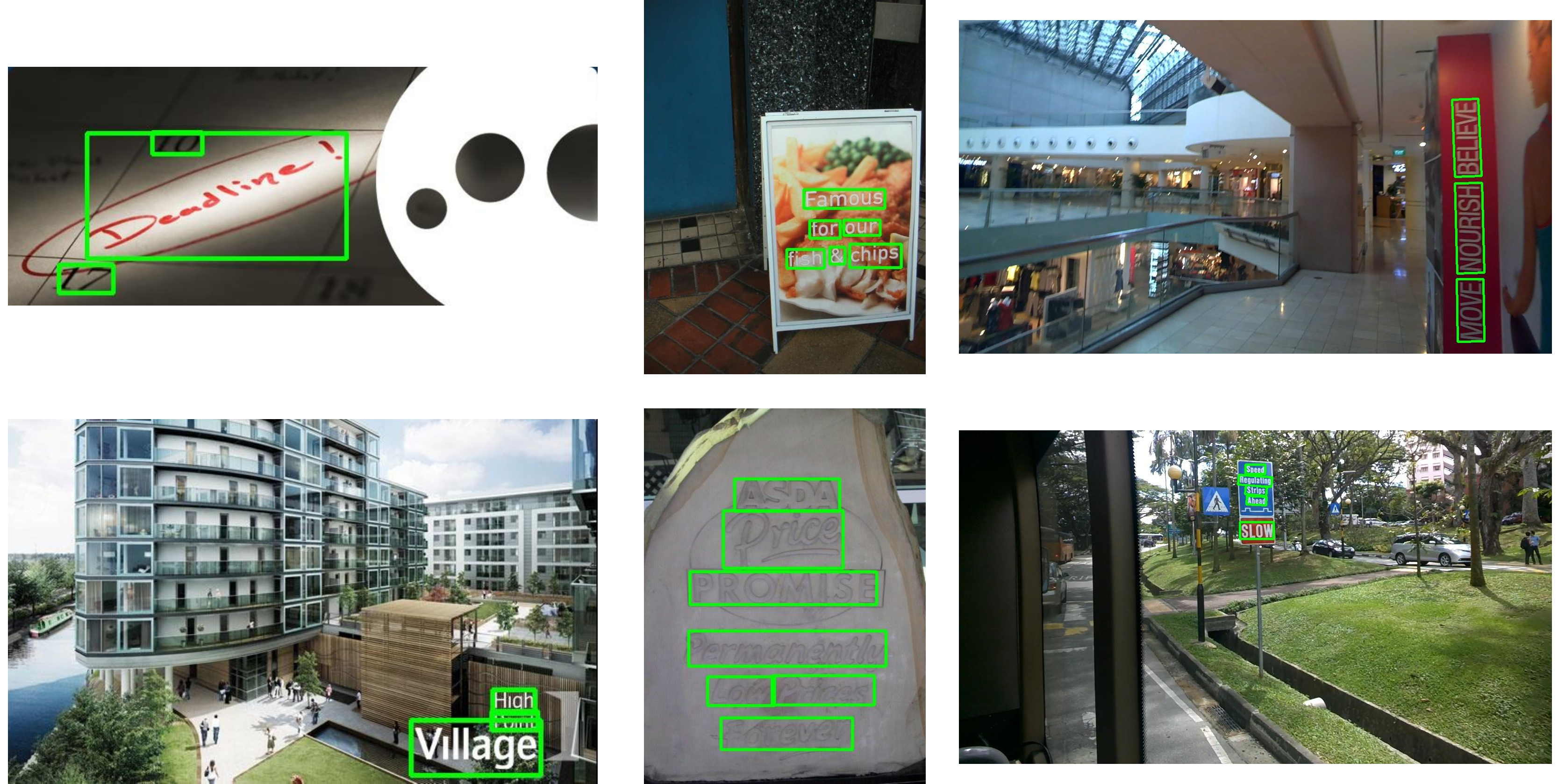}\label{fig:result-pos}} \quad
	\subfigure [Failure examples] {\includegraphics[width=0.34\linewidth, height=0.33\linewidth]{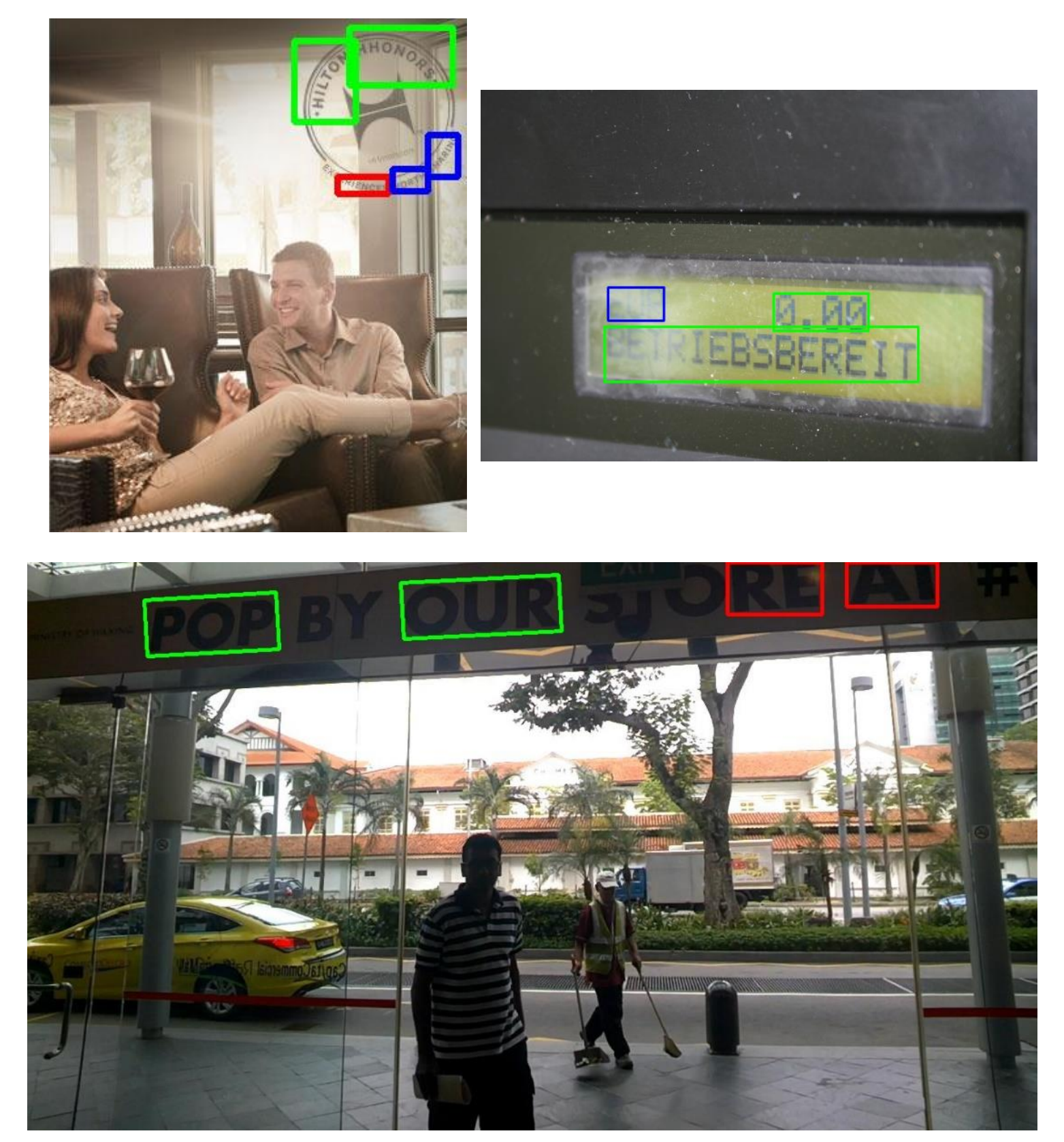}\label{fig:result-neg}}
	\caption{Examples of text detection results. Green bounding boxes: correct detections; Red solid boxes: false detections; Blue solid boxes: missed ground truths.}
	\label{fig:results}
\end{figure}

\begin{figure}[t]
	\centering
	\includegraphics[width=1.0\linewidth, height=0.3\linewidth]{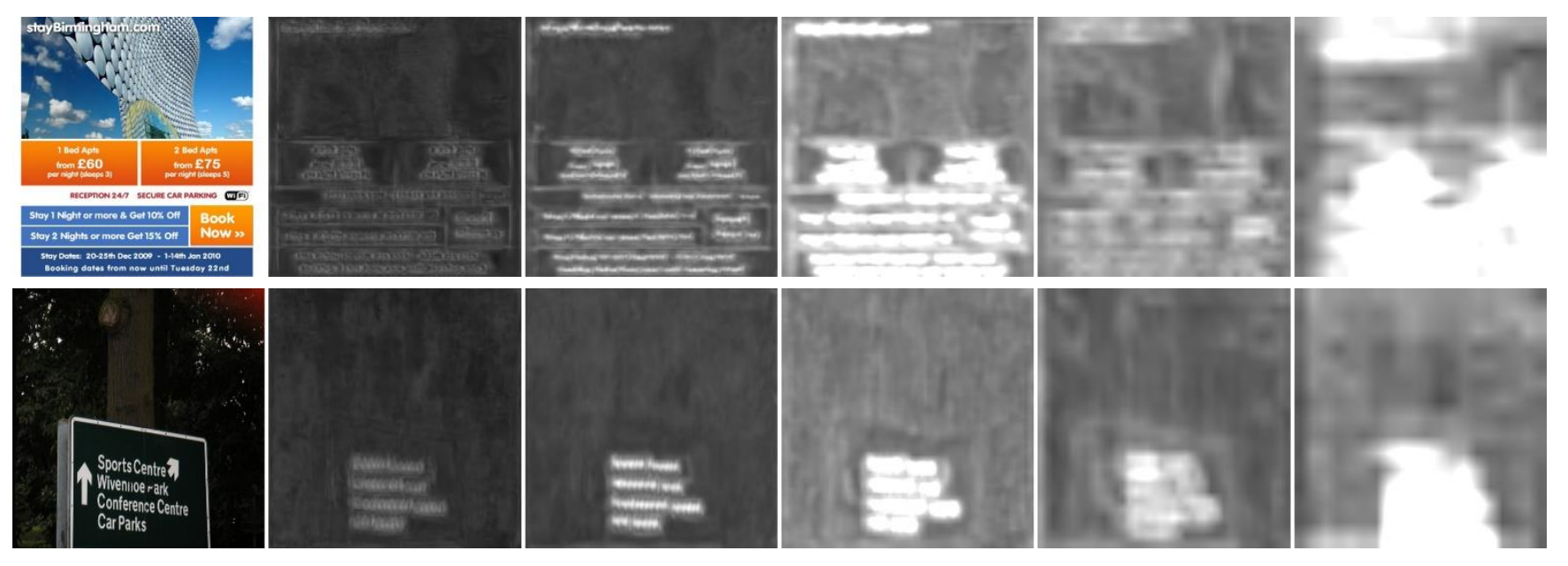}
	\caption{ The illustration of feature maps generated by BiFPN modules on ICDAR 2011 (top) and ICDAR 2013 (bottom). 1st column: Input image. 2nd-6th columns: The feature maps from level 1 ($l$=1) to level 5 ($l$=5).}
	\label{fig:feature-flow}
\end{figure}

\subsubsection{Qualitative Results}
\begin{figure}[t]
	\centering
	\subfigure [The weights of step 1] {\includegraphics[width=0.3\linewidth, height=0.26\linewidth]{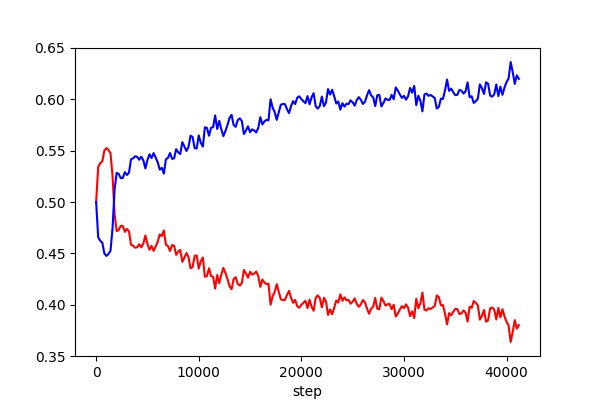}} \quad
	\subfigure [The weights of step 2] {\includegraphics[width=0.3\linewidth, height=0.26\linewidth]{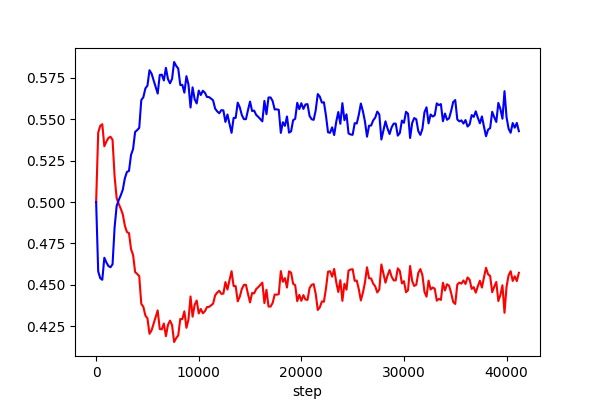}} \quad
	\subfigure [The weights of step 3] {\includegraphics[width=0.3\linewidth, height=0.26\linewidth]{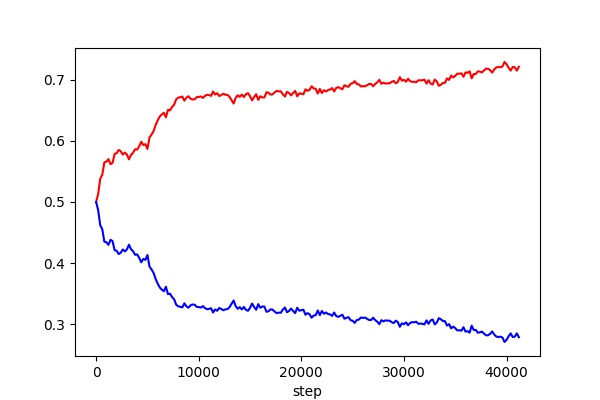}}
	\caption{The illustration of learned normalized weights for $f_1^{out}$ at three different BiFPN modules.
		The weights for input1 and input2 are represented in red and blue, respectively.}
	\label{fig:normed-weight}
\end{figure}

Several detection examples on ICDAR 2011, ICDAR 2013 and ICDAR 2015 datasets are shown in \figref{fig:results}. As shown, the results demonstrate that our proposed method has strong capability for detecting extremely ambiguous text lines and perspective distortion. 
However, our method still has a gap to achieve a perfect performance. Several failure cases are illustrated in \figref{fig:result-neg}. False positives and missing texts may appear in certain situations, such as non-uniform illumination, curved texts and occluded backgrounds.
The main contribution of our method is to enable the model to effectively extract robust features related to text instances owing to the multi-scale feature fusion module and the loss function.
In order to verify the effectiveness of our method, the extracted feature maps of each level are displayed in \figref{fig:feature-flow}. 
As can be seen in \figref{fig:feature-flow}, our model captures the local structures of text instances in the lower level layers and features with semantic information are extracted in the higher level layers.
Additionally, we illustrate the learned normalized weights for $f_1^{out}$ at three different BiFPN modules in \figref{fig:normed-weight} to further understand the behavior of Fast Normalized Fusion. As we expected, the normalized weights change during training which implies 
there exists suitable weights for the different features representing their contribution to the output.

\subsubsection{Quantitative Results}

We compared our proposed model with the previous state-of-the-art methods on three public benchmarks (ICDAR 2011, ICDAR 2013 and ICDAR 2015).
All experiments are performed with single image resolution that the longer side of the images in ICDAR 2011 and ICDAR 2013 are resized to 640 and that of ICDAR 2015 is resized to 1024. The image size are set differently since ICDAR 2015 has many small-scale texts. The performance of our method as well as other methods on three datasets are presented in \tabref{table:result}. Our method achieves state-of-the-art performance on ICDAR 2011 and ICDAR 2013 dataset, which demonstrates our approach extracts more robust features on text instances compared to other methods. In the case of ICDAR 2015, since our model trained only on ICDAR 2011 and ICDAR 2013 dataset due to the lack of character-level annotation in ICDAR 2015, the performance appear to be lower than other previous methods. However, when compared with the method such as Zhang et al. \cite{motd}, CTPN \cite{ctpn} and Yao et al. \cite{holistic}, our method is still competitive and leads the model to learn generalized feature representations for text detection. 

Furthermore, to highlight the effectiveness of the BiFPN module and the derived loss \eqref{eq:loss}, we finally perform an ablation study to compare the performances of our method with the following three models:
(1) Baseline : The model excluding both the BiFPN modules and the channel bottlenecks equipped with the regularization loss,
(2) Baseline + BiFPN : The model excluding only the channel bottlenecks, 
(3) Baseline + KL : The model excluding only the BiFPN modules. 
As shown in \tabref{table:result}, the performance results of all four models, including our proposed model, improve gradually from the baseline to our model in both recall and precision.
The remarkable performance improvement between the baseline and our proposed model,
especially in ICDAR 2015, implies that our method encourages the enrichment of feature representations on text instances.

\begin{table}[t]
	\caption{Results on three public benchmarks: ICDAR 2011, ICDAR 2013 and ICDAR 2015. 
		$R$, $P$ and $H$ represent recall, precision and H-mean, respectively. The best score is highlighted in \textbf{bold}. 
		The asterisk($^*$) denotes the results from the unseen dataset.}
	\label{table:result}
	\centering
	\begin{tabular}{c c c c||c c c||c c c}
		\toprule
		Method & \multicolumn{3}{c}{ICDAR 2011 \cite{icdar11}} & \multicolumn{3}{c}{ICDAR 2013 \cite{icdar13}} & \multicolumn{3}{c}{ICDAR 2015 \cite{icdar15}} \\
		& \multicolumn{3}{c}{(DetEval)} & \multicolumn{3}{c}{(DetEval)} &  \\
		& R & P & H & R & P & H & R & P & H \\
		\midrule
		Pal\textunderscore DAS16 \cite{pal_das16} & 87.95 & 91.14 & 89.51 & - & - & - & - & - & - \\ 
		SCUT\textunderscore HCII \cite{deeptext} & 81 & 85 & 83 & - & - & - & -  & - & - \\ 
		Text-CNN \cite{text-cnn} & 74 & 91 & 82 & 72.89 & 92.79 & 81.65 & - & - & - \\
		Ruan et al. \cite{fast} & 90.71 & 95.43 & 93.01 & 80.80 & 91.31 & 85.73 & 80.55 & 86.59 & 83.46 \\
		Zhang et al. \cite{motd} & - & - & - & 78 & 88 & 83 & 43 & 71 & 54 \\
		CTPN \cite{ctpn} & - & - & - & 83 & 93 & 88 & 52 & 74 & 61 \\ 
		Yao et al. \cite{holistic} & - & - & - & 80.2 & 88.8 & 84.3 & 58.7 & 72.3 & 64.8 \\ 
		SegLink \cite{seglink} & - & - & - & 83 & 87.7 & 85.3 & 76.8 & 73.1 & 75 \\ 		
		Texboxes++ \cite{textboxes++} & - & - & - & 86 & 92 & 89 & 78.5 & 87.8 & 82.9 \\ 
		RRPN \cite{rrpn} & - & - & - & 57.31 & 95.19 & 91.08 & 77.13 & 83.52 & 80.20 \\ 
		Stela \cite{stella} & - & - & - & 85.1 & 93.3 & 89 & 78.56 & 88.7 & 83.3 \\ 
		Mask Textspotter \cite{mask-textspotter} & - & - & - & 88.1 & 94.1 & 91 & 81.2 & 85.8 & 83.4 \\ 
		FOTS \cite{fots} & - & - & - & 90.47 & 94.63 & 92.5 & \textbf{85.17} & \textbf{91} & \textbf{87.99} \\ 
		CRAFT \cite{craft} & - & - & - & 93.06 & 97.43 & 95.20 & 84.26 & 89.79 & 86.93 \\ \hline
		\textbf{Baseline} & 92.68 & 94.24 & 93.45 & 93.15 & 97.50 & 95.27 & $53.64^*$ & $73.00^*$ & $61.84^*$ \\ 
		\textbf{Baseline + BiFPN} & 93.52 & 95.67 & 94.58 & 94.28 & 98.02 & 96.12 & $60.13^*$ & $68.36^*$ & $63.99^*$ \\ 
		\textbf{Baseline + KL} & 94.23 & 95.04 & 94.64 & 94.39 & 98.34 & 96.32 & $57.39^*$ & $71.98^*$ & $63.86^*$ \\ 
		\textbf{Proposed} & \textbf{94.35} & \textbf{95.90} & \textbf{95.12} & \textbf{95.14} & \textbf{98.48} & \textbf{96.78} & $61.24^*$ & $78.33^*$ & $68.74^*$ \\ 
		\bottomrule
	\end{tabular}
\end{table}

\section{Conclusion and Future Work}\label{sec:conclusion}
In this paper, we systemically designed a network architecture to maximize conditional log-likelihood and adopted a multi-scale feature fusion module to effectively aggregate the features generated from the backbone network. 
The proposed method achieves state-of-the-art performances on two public benchmarks (ICDAR 2011 and ICDAR 2013) and shows the generalization ability through ICDAR 2015 considering the unseen data. We believe that 
our proposed network architecture can be applied to all segmentation-based model and can lead to additional performance improvements. For the potential future work, we will apply the proposed network architecture to other previous models based on word-level annotation and continue to develop a better design of the networks.

\end{document}